%% file: main.tex
\documentclass[10pt,twocolumn,letterpaper]{article}
\usepackage[pagenumbers]{cvpr}

\input{math_commands.tex}

\usepackage{multirow}
\definecolor{cvprblue}{rgb}{0.21,0.49,0.74}
\usepackage[breaklinks,colorlinks,allcolors=cvprblue]{hyperref}
\usepackage[all]{hypcap}

\title{SMR: Learning Streaming Motion Retargeting from Human Motion Semantics}
\author{%
  Chun Gu\textsuperscript{1,2,3},
  Juntong Zhang\textsuperscript{2,3},
  Jingyu Li\textsuperscript{1,2,3},
  Li Zhang\textsuperscript{1,2,3$\dagger$}\\[0.4em]
  \normalsize
  \textsuperscript{1}School of Data Science, Fudan University\quad
  \textsuperscript{2}Shanghai Innovation Institute\quad
  \textsuperscript{3}Logos Intelligence Lab\\[0.4em]
  \normalsize\url{https://logosintelligencelab.github.io/SMR/}%
}
\hypersetup{
  pdftitle={SMR: Learning Streaming Motion Retargeting from Human Motion Semantics},
  pdfauthor={Chun Gu, Juntong Zhang, Jingyu Li, Li Zhang}
}

\begin{document}
\maketitle
\begingroup
  \renewcommand{\thefootnote}{\fnsymbol{footnote}}%
  \footnotetext[2]{Corresponding author: \href{mailto:lizhangfd@fudan.edu.cn}{\nolinkurl{lizhangfd@fudan.edu.cn}}.}
\endgroup

\suppressfloats[t]
\input{files/0-abstract}
\input{figures/teaser}
\input{files/1-introduction}
\input{figures/pipeline}
\input{files/2-relatedwork}
\input{files/3-method}
\input{files/4-experiment}
\input{files/5-conclusion}

\section*{LLM Usage Statement}
We used OpenAI Codex to assist with writing and revising project code and
to edit the manuscript for grammar, wording, and clarity.

{
  \small
  \bibliographystyle{ieeenat_fullname}
  \bibliography{main}
}

\clearpage
\appendix
\setcounter{table}{0}
\setcounter{figure}{0}
\setcounter{equation}{0}
\renewcommand{\thetable}{S\arabic{table}}
\renewcommand{\thefigure}{S\arabic{figure}}
\renewcommand{\theequation}{S\arabic{equation}}
\renewcommand{\theHtable}{appendix.\arabic{table}}
\renewcommand{\theHfigure}{appendix.\arabic{figure}}
\renewcommand{\theHequation}{appendix.\arabic{equation}}

\input{files/9-appendix}
\begin{samepage}
\input{files/7-ethics-statement}
\par
\end{samepage}
\begin{samepage}
\input{files/8-reproducibility-statement}
\par
\end{samepage}

\end{document}

%% file: math_commands.tex
\usepackage{amsmath,amsfonts,bm}

\def\eqref#1{equation~\ref{#1}}

\def\1{\bm{1}}

\DeclareMathAlphabet{\mathsfit}{\encodingdefault}{\sfdefault}{m}{sl}
\SetMathAlphabet{\mathsfit}{bold}{\encodingdefault}{\sfdefault}{bx}{n}



%% file: files/0-abstract.tex
\begin{abstract}
Retargeting human motion to humanoids requires preserving motion semantics
while accounting for the robot's physical constraints. Kinematic pipelines
prescribe a robot trajectory through geometric matching, without establishing
whether it can be executed under the robot's dynamics. Prior physics-supervised
neural methods train bidirectional full-sequence models on physics-refined
kinematic trajectories, remaining tied to prescribed kinematic solutions and
requiring noncausal context. We propose SMR, a streaming neural motion
retargeter learned from semantic-conditioned physics experts. We convert human
motion into robot-aware task-space targets for body landmark positions,
orientations, and velocities. Physics experts track these targets directly in
simulation under contact, dynamics, and actuator constraints. We train a single
strictly causal Transformer on their executed state trajectories that complete
the motion and pass semantic and physical validity checks. At each incoming
frame, the student uses only current and past human observations to generate a
robot motion reference for a downstream whole-body controller. Experiments
demonstrate that semantic-conditioned
physics experts generate motion with higher semantic fidelity and physical
validity than direct optimization-based retargeting, while the resulting
streaming retargeter achieves improved closed-loop performance over kinematic
and physics-refined baselines under a common whole-body controller.
\end{abstract}

%% file: figures/teaser.tex
\begin{figure}[t]
  \centering
  \includegraphics[width=\linewidth]{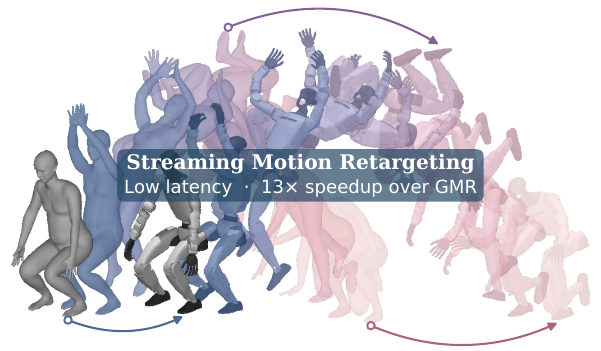}
  \caption{SMR streams one Unitree G1 reference per human input frame. Arrows
  connect human and robot poses at matching timestamps during a
  360\textdegree{} flip. The speedup compares SMR's CUDA Graph model step
  with GMR's iterative solve using the mean of three warm-run medians for
  each method (\cref{tab:inference_latency}).}
  \label{fig:teaser}
\end{figure}

%% file: files/1-introduction.tex
\section{Introduction}

Large-scale human motion datasets such as AMASS, LAFAN1, and BONES-SEED provide
a rich source of whole-body motion for humanoid learning and control
\citep{mahmood2019amass,harvey2020lafan1,studio2026bones}. Transferring these
motions to robots requires more than geometric matching because humans and
robots differ in morphology, joint structure, mass distribution, contact
mechanics, and actuation. For streaming applications such as live teleoperation,
retargeting must satisfy two requirements: preserving human motion semantics
in physically executable robot references, and generating those references from
current and past human observations.

The first requirement depends on how robot motion is generated for supervision.
Kinematic retargeting pipelines use inverse kinematics or geometric optimization
to construct robot trajectories
\citep{penco2018robust,darvish2019whole,araujo2025gmr}. Although these methods
can incorporate contact and collision constraints, geometric agreement alone
does not establish dynamic executability. Using the resulting trajectories as
neural supervision can tie the learned mapping to solutions selected by the
kinematic solver. Physics-supervised methods such as NMR~\citep{zhao2026nmr}
refine these trajectories through simulated tracking, but still use the
prescribed kinematic motion as the expert's tracking target. This motivates
specifying the motion to preserve directly through robot-aware task-space
semantics and using the physical execution to generate robot
supervision.

The second requirement is causal inference from a live human stream.
NMR's learned retargeter~\citep{zhao2026nmr} uses bidirectional modeling over a
complete motion sequence and therefore depends on future observations. This
prevents use in a streaming setting where the reference at time $t$ must
be generated from observations available up to $t$. The challenge is therefore
to retain the benefits of physics-based supervision while learning a causal
retargeter.

To address these challenges, we propose SMR, a streaming neural motion
retargeter learned from semantic-conditioned physics experts. We represent
human motion as robot-aware task-space targets for body landmark positions,
orientations, and velocities. The experts track these targets directly in
closed-loop simulation, using robot state and limited future semantic context
to generate motion under contact, dynamics, and actuator constraints. IK is
used only for initialization, rather than to prescribe a tracking trajectory.
We retain executed state rollouts that complete the motion and pass semantic
and physical validity checks, and use them to train a single strictly causal
Transformer. At each incoming frame, the student generates a robot motion
reference for a downstream whole-body controller using only current and past
human observations. \Cref{fig:teaser} shows the resulting Unitree G1
reference poses for a 360\textdegree{} flip.

Experiments show that semantic-conditioned physics experts generate motion
with higher semantic fidelity and physical validity than direct
optimization-based retargeting. Under the same frozen whole-body controller,
SMR achieves improved closed-loop performance over kinematic and physics-refined
baselines. We further demonstrate whole-body motion execution on a physical
Unitree G1 using SMR-generated references.

Our contributions are threefold: (1) we formulate physics-based retargeting around robot-aware human semantics, allowing experts to determine the robot trajectory instead of refining a prescribed kinematic solution; (2) we distill privileged expert rollouts into a single causal retargeter that generates streaming motion references from human observations, without robot-state inputs or future frames; and (3) we demonstrate improved closed-loop performance over kinematic and physics-refined baselines under a common frozen controller, alongside whole-body motion demonstrations on a physical Unitree G1.

%% file: figures/pipeline.tex
\begin{figure*}[t]
  \centering
  \includegraphics[width=\linewidth]{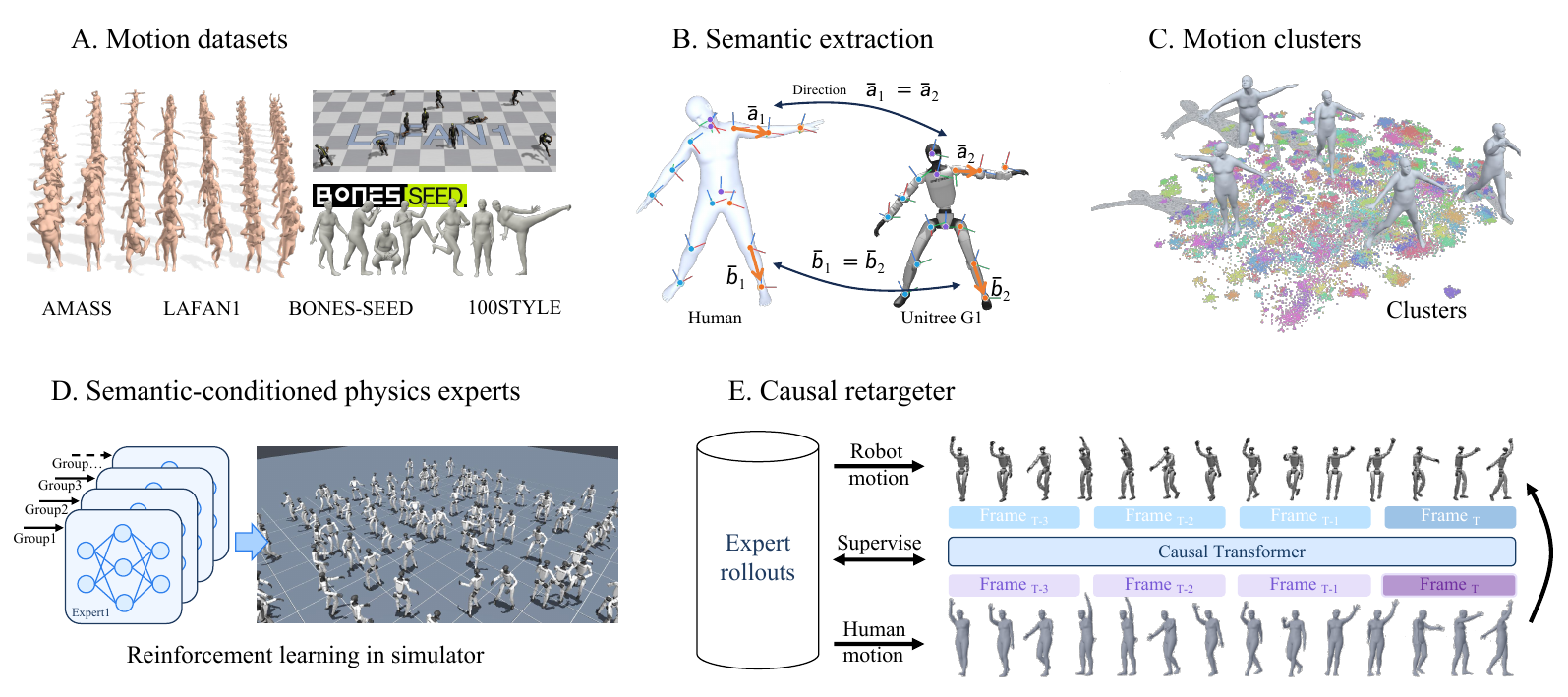}
  \caption{Overview of SMR. (A--C) Human motions from multiple datasets are
  converted into robot-aware task-space semantics and grouped for expert
  training. (D) Physics experts track these targets
  directly in simulation. (E) Accepted expert state rollouts supervise a
  single strictly causal transformer that generates robot motion references
  from current and past human observations.}
  \label{fig:pipeline}
\end{figure*}

%% file: files/2-relatedwork.tex
\section{Related work}

\subsection{Optimization-based motion retargeting}

Early motion retargeting methods transferred motion between articulated
characters through constrained optimization
\citep{gleicher1998retargeting,popovic1999physically,tak2005physically}.
For humanoids, geometric methods use Cartesian objectives to accommodate
differences in morphology and joint structure
\citep{penco2018robust,darvish2019whole}. Later approaches improve morphology
adaptation and geometric matching \citep{araujo2025gmr}, incorporate physical
constraints into data generation \citep{lee2026phuma}, and preserve human--object
or human--environment interactions \citep{yang2026omniretarget}. UMR further
uses learned surface correspondences to guide constrained pose matching and
contact transfer \citep{cao2026umr}. Dynamics-aware methods incorporate force
equilibrium \citep{rouxel2022multicontact}, optimal control
\citep{grandia2023doc}, or kinodynamic trajectory optimization
\citep{zhang2026kdmr,wang2026spark,dhedin2026dynaretarget}. Direct Dynamic
Retargeting uses sampling-based model predictive control to optimize task-space
imitation directly in simulation, without an intermediate kinematic trajectory
\citep{roux2026ddr}. These methods generate robot motion by optimizing input
poses or trajectories. SMR instead learns from accepted physics-expert rollouts
and generates streaming references with a causal neural model, without
retargeting optimization at deployment.

\subsection{Learning-based motion retargeting}

Learning-based retargeters learn cross-skeleton mappings through shared latent
spaces, skeleton-aware architectures, and paired or unpaired motion
translation \citep{villegas2018neural,aberman2020skeleton,lee2023same,
choi2020nonparametric,yan2023imitationnet}. Online upper-body retargeting has
also been demonstrated using unpaired adversarial learning
\citep{yagi2024unsupervised}. Physics enters these models through feasibility
constraints in Human2Humanoid~\citep{huang2026human2humanoid} and simulated
tracking feedback in IKMR~\citep{chen2026ikmr}. ReActor uses sparse semantic
correspondences to jointly adapt reference motions and train a tracking policy
through bilevel optimization
\citep{muller2026reactor}, while OmniTrack uses privileged simulation rollouts
to generate references for training a downstream tracking policy
\citep{li2026omnitrack}. Like SMR, NMR uses rollouts from clustered RL experts
to train a neural retargeter. Its experts track GMR-generated trajectories, and
its retargeter processes complete motion sequences \citep{zhao2026nmr}. SMR
combines direct tracking of robot-aware human semantics in simulation with
distillation of accepted state rollouts into a strictly causal retargeter.

\subsection{Physics-based motion tracking}

Physics-based motion tracking studies how robots and simulated characters
execute reference motions under dynamics and contact constraints. DeepMimic
and PHC address this problem through reinforcement learning in simulation
\citep{peng2018deepmimic,luo2023phc}. Subsequent humanoid systems improve
sim-to-real transfer \citep{he2025asap} and motion diversity
\citep{liao2026beyondmimic,chen2025gmt,han2025kungfubot2}, introduce online
dynamics adaptation \citep{zhang2025any2track}, and scale motion tracking
through larger datasets and models \citep{luo2026sonic,qi2026humanoidgpt}.
MimicLite focuses on efficient training of general motion-tracking policies
\citep{robopartylab2026mimiclite}. These approaches use robot-state feedback
to produce control actions, whereas SMR generates motion references from human
observations. We evaluate these references with a frozen whole-body controller
to isolate the effect of retargeting on downstream execution.

%% file: files/3-method.tex
\section{Method}
\label{sec:method}

\subsection{Overview}
\label{sec:method_overview}

SMR maps a stream of human poses to joint and pelvis references for a
Unitree G1. To generate training data for the retargeter, we convert each
human motion into robot-aware landmark targets and train physics experts to
track them in simulation (\cref{sec:physics_expert}). We retain the expert
state trajectories that satisfy tracking and physical validity checks. A
single causal Transformer learns to predict these states from the human pose stream
(\cref{sec:neural_retargeting}). At deployment, the Transformer generates
references for a separate whole-body controller. \Cref{fig:pipeline}
summarizes training and inference.

\subsection{Semantic-conditioned physics experts}
\label{sec:physics_expert}

\paragraph{Robot-aware motion targets.}
We represent motions from AMASS~\citep{mahmood2019amass},
LAFAN1~\citep{harvey2020lafan1}, BONES-SEED~\citep{studio2026bones},
100STYLE~\citep{mason2022realtime}, and smaller collections using neutral
SMPL-H at 50~Hz~\citep{loper2015smpl,romero2017embodiedhands}. Given a human
pose $H_t$, we run SMPL-H forward kinematics and express the skeleton in
the robot's Z-up world frame. We scale the root translation by the ratio
of robot to human neutral leg lengths, with a fixed vertical alignment
from the rest poses. We place the remaining landmarks along the human
segment directions using the robot's rest lengths and express the human
orientations in the corresponding robot landmark frames. For target robot
$\mathcal R$, the resulting command is
\begin{equation}
  c_t = \Psi_{\mathcal R}(H_t)
  = \left\{p_{k,t},R_{k,t},v_{k,t},\omega_{k,t}\right\}_{k\in\mathcal K},
  \label{eq:semantic_command}
\end{equation}
where $\mathcal K$ contains 15 landmarks, $p$ and $R$ are the transported
world-frame positions and orientations, and $v$ and $\omega$ are obtained
by finite differences after transport. These commands specify the
task-space motion to preserve. The physics expert determines the joint
trajectory. \Cref{fig:robot-aware-semantics}
shows the transported targets overlaid on the simulated robot.

\paragraph{Motion grouping.}
We train one expert per motion for the smaller collections. For
100STYLE and BONES-SEED, we use Harmonic Motion Embedding
(HME)~\citep{qi2026humanoidgpt} to group motions before expert training.
We fit a Periodic Autoencoder~\citep{starke2022deepphase} for each collection
on two-second windows of the robot-aware semantic stream. For each motion,
we summarize the 64 amplitude and 64 frequency channels by their means
and standard deviations across windows. We standardize these 256-D
descriptors and apply MiniBatch K-means with 32 clusters for 100STYLE
and 256 for BONES-SEED. Each cluster defines one expert's training pool.

\input{figures/robot-aware-semantics}

\paragraph{Expert training and initialization.}
We train each expert with proximal policy optimization
(PPO)~\citep{schulman2017proximal}. At control step $t$, the actor takes
\begin{equation}
  o_t = \operatorname{concat}\!\left(\tilde c_t,
    x_t^{\mathrm{hist}},a_{t-3:t-1}\right),
  \;
  a_t \sim \pi_e(\cdot\mid o_t),
  \label{eq:expert_policy}
\end{equation}
where $\tilde c_t$ is a heading-local window of command positions and
orientations, $x_t^{\mathrm{hist}}$ is sampled robot proprioception, and
$a_{t-3:t-1}$ contains the three preceding actions. The offline actor can
see up to 80~ms of future commands and outputs 29 joint-position targets
at 50~Hz. The critic and tracking rewards also use command velocities.
Root tracking errors are computed in world coordinates and body errors
in heading-local coordinates. The reward combines these tracking terms
with action and joint regularization, a self-collision penalty, and a
survival term. Appendix~A.3 lists the actor observations and reward weights.

At reset, we initialize the robot from a pose solved offline by
differential inverse kinematics on the same semantic targets. The solve
prioritizes landmark positions, uses softer orientation targets and a
weak previous-pose prior, and enforces joint and joint-step limits. The
resulting trajectory supplies reset states only. During the episode PPO
receives rewards for tracking $c_t$ directly, so the policy can realize
the same task with a different joint trajectory.

\paragraph{State rollout supervision.}
We replay each motion from its first frame with the assigned expert's
mean policy and keep the resulting trajectory if it reaches the final
frame and passes the tracking, joint-limit, speed, effort, and
self-collision checks specified in Appendix~A.4. The labels are the
recorded joint, root, and body states, aligned to the commands at 50~Hz.
\Cref{tab:expert_assignment} reports the expert assignments and screening
results for each motion source. Across 428 experts, 89{,}686 of 145{,}933
candidate rollouts pass all checks. We then split the accepted rollouts
by original motion group, keeping chunks and mirrored variants together.
The training split contains 80{,}728 rollouts and supervises the SMR checkpoint
used in \cref{fig:challenging_references,fig:qualitative_comparison}.

\input{tables/expert-assignment}

\subsection{Neural retargeting}
\label{sec:neural_retargeting}

\paragraph{Causal input and pelvis-relative output.}
At frame $t$ the student receives an 85-D input $u_t$ computed from the
human semantic stream. The 79-D pose comprises the 14 pelvis-relative
landmark positions (42-D), pelvis height (1-D), pelvis tilt with heading
removed (6-D), and 6-D orientations for the torso, wrists, and ankles
(30-D). Positions and orientations use the current human heading frame.
The 6-D pelvis twist uses backward differences at the observation rate
in the previous human pelvis frame and is zero at the first frame.
The student predicts
\begin{equation}
  \widehat m_t = f_\theta(u_{0:t})
  = \operatorname{concat}\!\left(\widehat q_t,\widehat{\Delta p}_t,
    \widehat{\Delta r}_t\right) \in \mathbb R^{35},
  \label{eq:causal_mapping}
\end{equation}
where $\widehat q_t$ contains the 29 G1 joint positions and
$\widehat{\Delta p}_t,\widehat{\Delta r}_t\in\mathbb R^3$ are the pelvis
translation and rotation-vector offsets. The reference pelvis pose
composes these offsets with the current transported human pelvis pose.
This anchors each output without integrating predicted root velocity.

\paragraph{Streaming Transformer.}
We use four pre-normalized blocks of width 512, 16 attention heads,
and rotary position embeddings~\citep{vaswani2017attention,su2024roformer}.
Each layer attends to the current token and up to 127 preceding tokens.
We scale rotary indices by $50/f$ at observation rate $f$ to encode elapsed
time at 50-Hz units while keeping the per-layer token budget fixed. At
inference each layer updates a circular key-value cache for each arriving
frame. Features and attention are causal.

\input{figures/challenging-references}
\input{tables/reference-quality}

\paragraph{Distillation.}
We form training windows at 30, 50, and 90~Hz by pairing resampled human
poses and expert states at the same timestamps. The student learns the
relative transform from the input pelvis to the expert pelvis, together
with the expert joint positions. During augmentation, the input pose
history receives temporally correlated noise and short holds. We
recompute input features and offsets against the same clean expert
states. The objective is
\begin{equation}
  \mathcal L = \mathcal L_{\mathrm{dist}}
  + 0.1\mathcal L_v
  + 0.01\mathcal L_{\mathrm{pen}}
  + 0.1\mathcal L_{\mathrm{slip}}.
  \label{eq:distillation}
\end{equation}
The distillation term is the mean squared error on normalized outputs.
$\mathcal L_v$ matches joint rates and reconstructed pelvis motion.
Differentiable forward kinematics supplies the foot geometry for
$\mathcal L_{\mathrm{pen}}$, which penalizes ground penetration, and
$\mathcal L_{\mathrm{slip}}$, which matches horizontal foot velocities
over support intervals inferred from the clean expert states.

%% file: figures/robot-aware-semantics.tex
\begin{figure*}[!t]
  \centering
  \includegraphics[width=\linewidth]{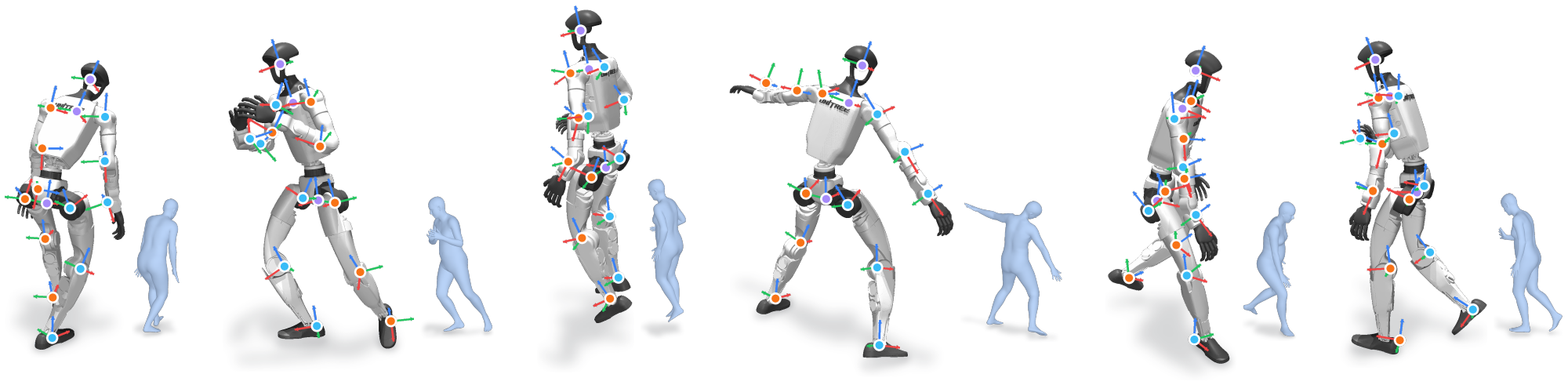}
  \caption{Robot-aware semantic tracking across six motions. Each inset shows
  the source SMPL-H pose beside the corresponding Unitree G1 pose produced by
  a physics expert in simulation. Colored markers and axes show the transported
  landmark targets.}
  \label{fig:robot-aware-semantics}
\end{figure*}

%% file: tables/expert-assignment.tex
\begin{table}[!t]
  \centering
  \footnotesize
  \setlength{\tabcolsep}{2pt}
  \caption{Unitree G1 expert assignments and rollout acceptance by motion source.}
  \label{tab:expert_assignment}
  \begin{tabular*}{\columnwidth}{@{\extracolsep{\fill}}llrrr@{}}
    \toprule
    Source & Assignment & Experts & Candidates & Accepted \\
    \midrule
    BONES-SEED~\citep{studio2026bones} & Clustered & 256 & 142{,}220 & 86{,}134 \\
    100STYLE~\citep{mason2022realtime} & Clustered & 32 & 3{,}573 & 3{,}460 \\
    LAFAN1~\citep{harvey2020lafan1} & Single-motion & 40 & 40 & 26 \\
    AMASS~\citep{mahmood2019amass} & Single-motion & 22 & 22 & 10 \\
    Other (5 sets) & Single-motion & 78 & 78 & 56 \\
    \midrule
    Total & & 428 & 145{,}933 & 89{,}686 \\
    \bottomrule
  \end{tabular*}
\end{table}

%% file: figures/challenging-references.tex
\begin{figure*}[!t]
  \centering
  \includegraphics[width=\linewidth]{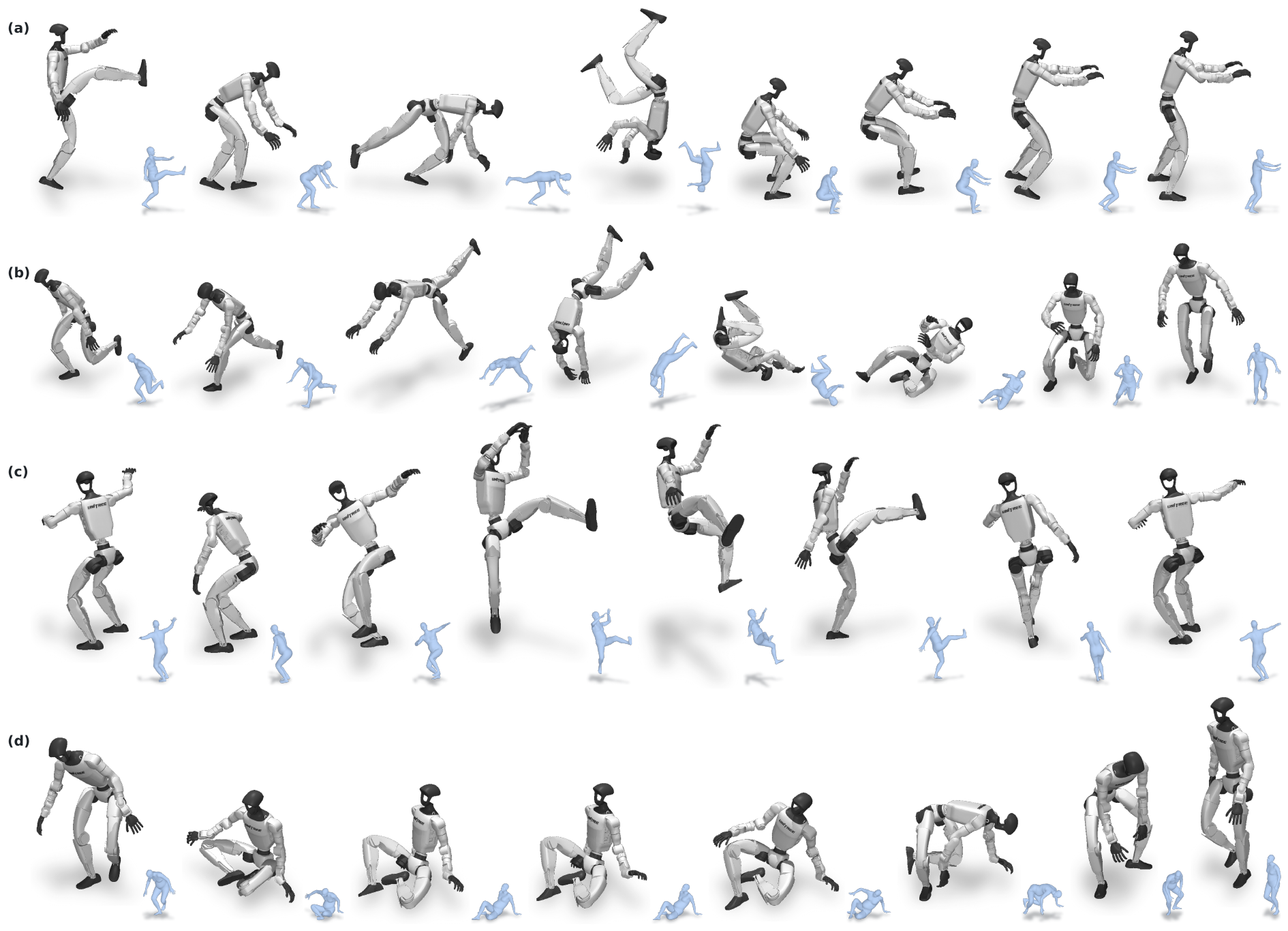}
  \caption{SMR references for challenging motions, with temporally matched
  SMPL-H inputs in blue. (a) Webster flip, (b) jump into a shoulder roll,
  (c) spinning kick, and (d) fall and get up. Inputs in (a)--(c) were
  reconstructed from Unitree G1 trajectories. The input in (d) comes from
  LaFAN1~\citep{harvey2020lafan1}.}
  \label{fig:challenging_references}
\end{figure*}

%% file: tables/reference-quality.tex
\begin{table*}[t]
  \centering
  \scriptsize
  \setlength{\tabcolsep}{1.5pt}
  \renewcommand{\arraystretch}{1.08}
  \caption{Test-set reference quality.
  FK and Sem. are world-frame and root-relative landmark errors. Root and End.
  are mean and final pelvis position errors. Limit and Coll. give interval rates
  of joint-limit violations and self-collisions.
  Slip is support-foot sliding speed.}
  \label{tab:reference_quality}
  \begin{tabular*}{\textwidth}{@{\extracolsep{\fill}}lccccccc}
    \toprule
    Reference source
    & FK (m)$\,\downarrow$
    & Sem. (m)$\,\downarrow$
    & Root (m)$\,\downarrow$
    & End. (m)$\,\downarrow$
    & Limit (\%)$\,\downarrow$
    & Coll. (\%)$\,\downarrow$
    & Slip (m/s)$\,\downarrow$ \\
    \midrule
    GMR & 0.077 & 0.029 & 0.064 & 0.075 & \textbf{0.00} & 3.53 & 0.045 \\
    NMR & 0.163 & 0.032 & 0.156 & 0.387 & \textbf{0.00} & 2.75 & 0.050 \\
    GMR-supervised Student & 0.063 & 0.030 & 0.051 & 0.044 & 29.59 & 2.90 & 0.040 \\
    GMR-Expert Student & 0.068 & 0.034 & 0.053 & 0.045 & 0.08 & \textbf{1.78} & 0.033 \\
    SMR (Ours) & \textbf{0.027} & \textbf{0.020} & \textbf{0.016} & \textbf{0.019} & 0.01 & 1.79 & \textbf{0.030} \\
    \midrule
    GMR-conditioned Expert & 0.082 & 0.038 & 0.064 & 0.075 & \textbf{0.02} & \textbf{0.01} & 0.024 \\
    Semantic Expert (Ours) & \textbf{0.028} & \textbf{0.023} & \textbf{0.013} & \textbf{0.011} & \textbf{0.02} & \textbf{0.01} & \textbf{0.017} \\
    \bottomrule
  \end{tabular*}
\end{table*}

%% file: files/4-experiment.tex
\section{Experiments}
\label{sec:experiments}

\input{figures/qualitative-comparison}
\input{tables/closed-loop-matrix}

To assess whether SMR preserves human motion semantics while producing
references the robot can execute, we evaluate it on the Unitree G1. We
examine reference quality and closed-loop execution, alongside further
analyses and qualitative streaming teleoperation on hardware. Implementation
details are provided in Appendices~\ref{sec:supp_physics_experts} and~\ref{sec:supp_neural_retargeter}.

\input{figures/scaling}
\input{tables/objective-augmentation}
\input{tables/inference-latency}

\subsection{Expert and student reference quality}
\label{sec:expert_student_experiments}

We ask whether tracking robot-aware task-space semantics yields more
faithful references than tracking a prescribed GMR trajectory, and
whether SMR retains that fidelity after causal distillation
(\Cref{tab:reference_quality}). The GMR-conditioned RL expert tracks
GMR-derived trajectories for each motion cluster, and its accepted state
rollouts train the GMR-Expert Student. The GMR-supervised Student learns
from GMR references, while SMR learns from accepted semantic-expert
rollouts. We evaluate all reference sources on the same 4{,}658 motions.
We use NMR's released checkpoint with full-sequence inference. The SMR entries in
\cref{tab:reference_quality,tab:closed_loop_matrix} come from the original
Student-Final checkpoint.
Physics-based tracking of GMR references greatly reduces self-collisions,
while landmark error remains close to GMR's. The semantic expert lowers
world-frame landmark error from 8.2~cm to 2.8~cm compared with the
GMR-conditioned expert. Both experts report a self-collision rate of
0.01\%. Root-relative landmark error drops from 3.8 to 2.3~cm, indicating
closer agreement in body configuration around the pelvis. The SMR student
has a root-relative error of 2.0~cm, compared with 2.3~cm for its expert
and 3.0--3.4~cm for the GMR-based students. After distillation, the self-collision rate
rises to 1.79\%, and support-foot slip increases from 0.017 to 0.030~m/s.
We train the experts in simulation with a self-collision penalty and screen
their rollouts for collisions. At inference, SMR predicts robot poses without
imposing hard contact constraints. Small pose errors can therefore cause
nearby links to intersect or a support foot to slide.
SMR also generates references for challenging motions involving aerial rotation
and transitions between standing and ground contact
(\Cref{fig:challenging_references}). The references capture inversion in the
Webster flip, the transition from a jump into a shoulder roll, leg extension
during a spinning kick, and the return to standing after a fall.
These examples use the SMR checkpoint trained for 50,000 steps on the expert
rollouts summarized in \cref{tab:expert_assignment}.
\Cref{fig:qualitative_comparison} compares poses at matched timestamps for
four motions. SMR more closely preserves the source arm alignment during
the overhead reach and captures both knee flexion and torso alignment in
the half squat.

\subsection{Closed-loop execution}
\label{sec:closed_loop_experiments}

We next evaluate whether the generated references can be executed while
preserving the intended human motion. \Cref{tab:closed_loop_matrix}
compares five methods on the same 400 motions with a frozen MimicLite controller in
mjlab~\citep{zakka2026mjlab}. All 400 motions were in the expert training pools.
Their source motion groups were held out during student training.
The Unitree G1 asset, scene, actuators,
initialization, and reference offsets are fixed across methods. Success
requires completion without absolute pelvis-height error above 0.30~m or
an absolute projected-gravity vertical-component difference above 0.8.
Non-finite or incomplete rollouts fail. MPJPE measures tracking of the robot
reference, and E-MPJPE measures how closely the executed wrists and ankles follow the
human targets. Both errors include all 400 motions, including failures.
The benchmark SMR checkpoint completes 97.0\% of
motions, compared with 90.0\% for GMR, 92.0\% for NMR, and 91.5--93.0\%
for the GMR-based students. It also achieves the lowest MPJPE (0.039~m)
and E-MPJPE (0.080~m). Together, higher completion and lower execution
errors show that SMR's streaming references, distilled from
dynamics-grounded expert rollouts, serve as effective targets for the
frozen WBC to drive Unitree G1 in simulation while retaining the intended human
motion.

\subsection{Ablations}
\label{sec:ablation_scaling_experiments}

\input{figures/real-robot-teleop}

To assess the effects of data and model scale, we train the Base student
with different fractions of the training set and different widths
(\Cref{fig:scaling}). Held-out error decreases markedly up to 50\%
of the training data, but doubling the data to 100\% brings little further
gain. Under the same training budget, increasing the model beyond 12.7M
parameters slightly worsens held-out error despite lower training loss,
suggesting that the added capacity may overfit at this data scale. We
use the 12.7M-parameter architecture in both SMR checkpoints.
\Cref{fig:controller_training_curves} shows a portion of the MimicLite
training curves for GMR, NMR, and SMR references.
Each method uses one run with 8,192 environments per GPU on eight GPUs.
At the end of the displayed window, SMR has the longest episodes, while
NMR has the highest mean reward. The curves describe training on each
method's reference distribution, with method-specific reference processing.
\Cref{tab:objective_augmentation} compares students trained with
and without auxiliary losses and input augmentation. The clearest effect
of the auxiliary losses is on support-foot slip, which drops from
0.0512 to 0.0325~m/s when the losses are added to the Base student.
The reduction is similar with augmentation enabled, from 0.0501 to
0.0329~m/s.
Input augmentation lowers both closed-loop tracking errors with either loss
configuration. In the model with auxiliary losses, augmentation reduces
E-MPJPE from 0.0814 to 0.0795~m even as the root-relative reference error
increases slightly. Using both gives the lowest closed-loop errors.
Completion varies by at most 0.25 percentage points across the four variants.

\subsection{Streaming real-robot deployment}
\label{sec:streaming_experiments}

In the 50~Hz retargeting benchmark, SMR's model step takes 0.37~ms per
frame and GMR's iterative solve takes 4.87~ms
(\Cref{tab:inference_latency}). The values average three warm-run medians.
SMR also achieves higher completion and lower execution errors under
the frozen WBC (\Cref{tab:closed_loop_matrix}), combining faster
reference generation with better closed-loop simulation performance.
\Cref{fig:real_robot_teleop} shows streaming deployment on a
physical Unitree G1. In both cases, a MimicLite policy controls the robot using
motion references from the retargeter. GMR's kinematic optimization does
not enforce dynamically stable execution: the GMR-driven robot becomes
unstable while kneeling, and its raised-arm motion falls out of phase
with the operator. With SMR references, MimicLite produces smooth,
stable motion throughout the demonstrated teleoperation sequences.

%% file: figures/qualitative-comparison.tex
\begin{figure*}[t]
  \centering
  \includegraphics[width=\linewidth]{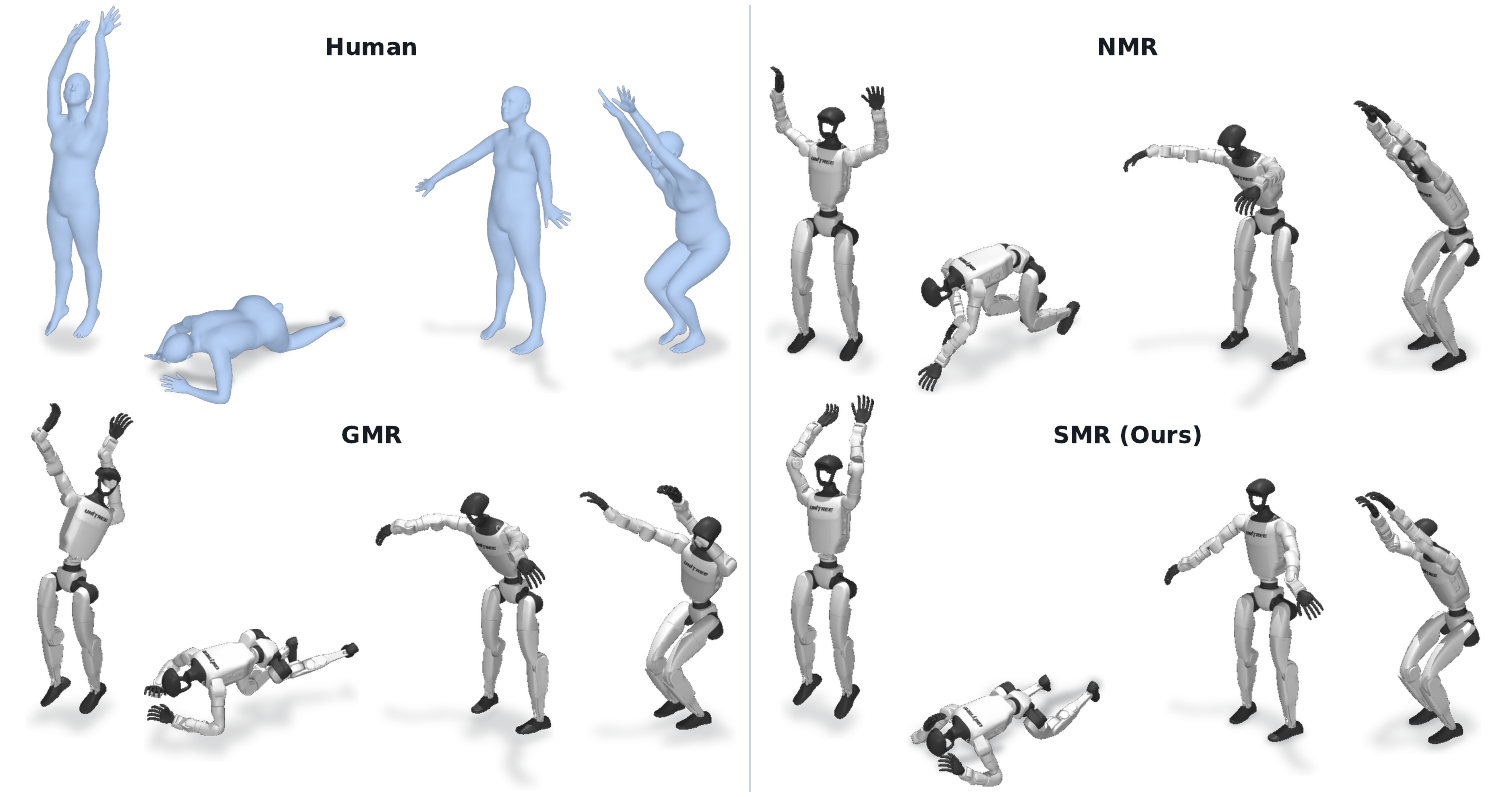}
  \caption{Selected examples of source posture preservation on Unitree G1.
  The same four motions appear in both blocks, with source SMPL-H and GMR
  on the left and NMR and SMR on the right. Each motion is shown at a
  matched timestamp. For display, human poses are scaled to match the
  neutral height of the G1. SMR uses the 50{,}000-step checkpoint from
  \cref{fig:challenging_references}.}
  \label{fig:qualitative_comparison}
\end{figure*}

%% file: tables/closed-loop-matrix.tex
\begin{table*}[t]
  \centering
  \scriptsize
  \setlength{\tabcolsep}{3pt}
  \renewcommand{\arraystretch}{1.08}
  \caption{Closed-loop evaluation on 400 motions using the same frozen whole-body controller. SMR achieves the highest success rate and lowest tracking and human-target errors, demonstrating more reliable execution and better preservation of human motion semantics.}
  \label{tab:closed_loop_matrix}
  \begin{tabular*}{0.98\textwidth}{@{\extracolsep{\fill}}lccccc}
    \toprule
    & GMR
    & NMR
    & GMR-supervised Student
    & GMR-Expert Student
    & SMR (Ours) \\
    \midrule
    Success (\%)$\,\uparrow$ & 90.0 & 92.0 & 91.5 & 93.0 & \textbf{97.0} \\
    MPJPE (m)$\,\downarrow$ & 0.060 & 0.057 & 0.055 & 0.051 & \textbf{0.039} \\
    E-MPJPE (m)$\,\downarrow$ & 0.115 & 0.111 & 0.111 & 0.112 & \textbf{0.080} \\
    \bottomrule
  \end{tabular*}
\end{table*}

%% file: figures/scaling.tex
\begin{figure*}[t]
  \centering
  \begin{minipage}[t]{0.49\textwidth}
    \centering
    \includegraphics[width=\linewidth]{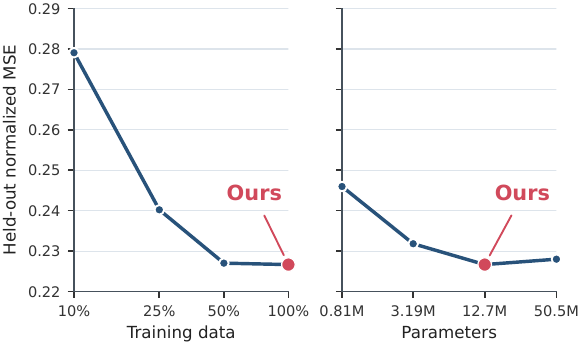}
    \captionof{figure}{Data and model scaling. Gains diminish beyond 50\% of the data, and the 12.7M model has the lowest held-out error.}
    \label{fig:scaling}
  \end{minipage}\hfill
  \begin{minipage}[t]{0.49\textwidth}
    \centering
    \includegraphics[width=\linewidth]{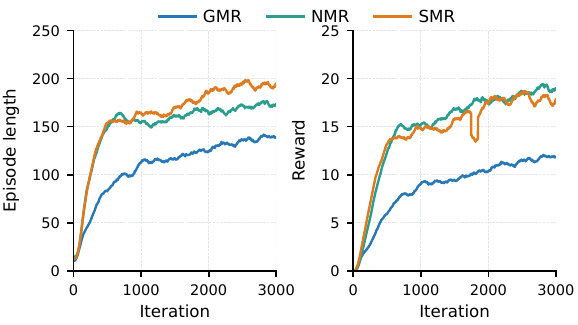}
    \captionof{figure}{MimicLite training by reference source.}
    \label{fig:controller_training_curves}
  \end{minipage}
\end{figure*}

%% file: tables/objective-augmentation.tex
\begin{table*}[!t]
  \centering
  \scriptsize
  \setlength{\tabcolsep}{2.5pt}
  \renewcommand{\arraystretch}{1.08}
  \caption{Matched ablation of auxiliary losses (Aux.: velocity, foot
  penetration, support-foot slip) and input augmentation (Aug.: correlated
  pose noise and 1--2-frame causal holds). Sem. and Slip use the metrics
  defined in \Cref{tab:reference_quality}. MPJPE, E-MPJPE, and Success
  are measured with the frozen WBC.
  Lower is better for errors, higher for Success. Best values are bold.}
  \label{tab:objective_augmentation}
  \begin{tabular*}{\textwidth}{@{\extracolsep{\fill}}lcccccccc}
    \toprule
    Variant & Aux. & Aug.
    & MSE (norm.)$\,\downarrow$
    & Sem. (m)$\,\downarrow$
    & Slip (m/s)$\,\downarrow$
    & MPJPE (m)$\,\downarrow$
    & E-MPJPE (m)$\,\downarrow$
    & Success (\%)$\,\uparrow$ \\
    \midrule
    Base & --- & --- & 0.24580 & 0.0288 & 0.0512 & 0.0428 & 0.0839 & 96.75 \\
    Aux-only & \checkmark & --- & \textbf{0.24521} & \textbf{0.0276} & \textbf{0.0325} & 0.0412 & 0.0814 & 96.75 \\
    Aug-only & --- & \checkmark & 0.24642 & 0.0287 & 0.0501 & 0.0411 & 0.0819 & 96.75 \\
    SMR (Ours) & \checkmark & \checkmark & 0.24863 & 0.0283 & 0.0329 & \textbf{0.0392} & \textbf{0.0795} & \textbf{97.00} \\
    \bottomrule
  \end{tabular*}
\end{table*}

%% file: tables/inference-latency.tex
\begin{table}[!t]
  \centering
  \scriptsize
  \setlength{\tabcolsep}{2pt}
  \renewcommand{\arraystretch}{1.08}
  \caption{Per-frame latency of GMR~\citep{araujo2025gmr} and
  four SMR model sizes. Compared with Small and Medium, Large has substantially
  more parameters but only slightly higher latency. Together with its lowest
  held-out error in \Cref{fig:scaling}, this makes Large a favorable
  balance of accuracy and speed. Bold marks the lowest latency.}
  \label{tab:inference_latency}
  \begin{tabular*}{\columnwidth}{@{\extracolsep{\fill}}lccccc@{}}
    \toprule
    & \multirow{2}{*}{GMR} & \multicolumn{4}{c@{}}{SMR} \\
    & & Small & Medium & Large (Ours) & XLarge \\
    \midrule
    Parameters (M) & -- & 0.81 & 3.19 & 12.7 & 50.5 \\
    Median (ms/frame)$\,\downarrow$ & 4.87 & \textbf{0.34} & 0.35 & 0.37 & 0.58 \\
    \bottomrule
  \end{tabular*}
\end{table}

%% file: figures/real-robot-teleop.tex
\begin{figure*}[!t]
  \centering
  \includegraphics[width=\linewidth]{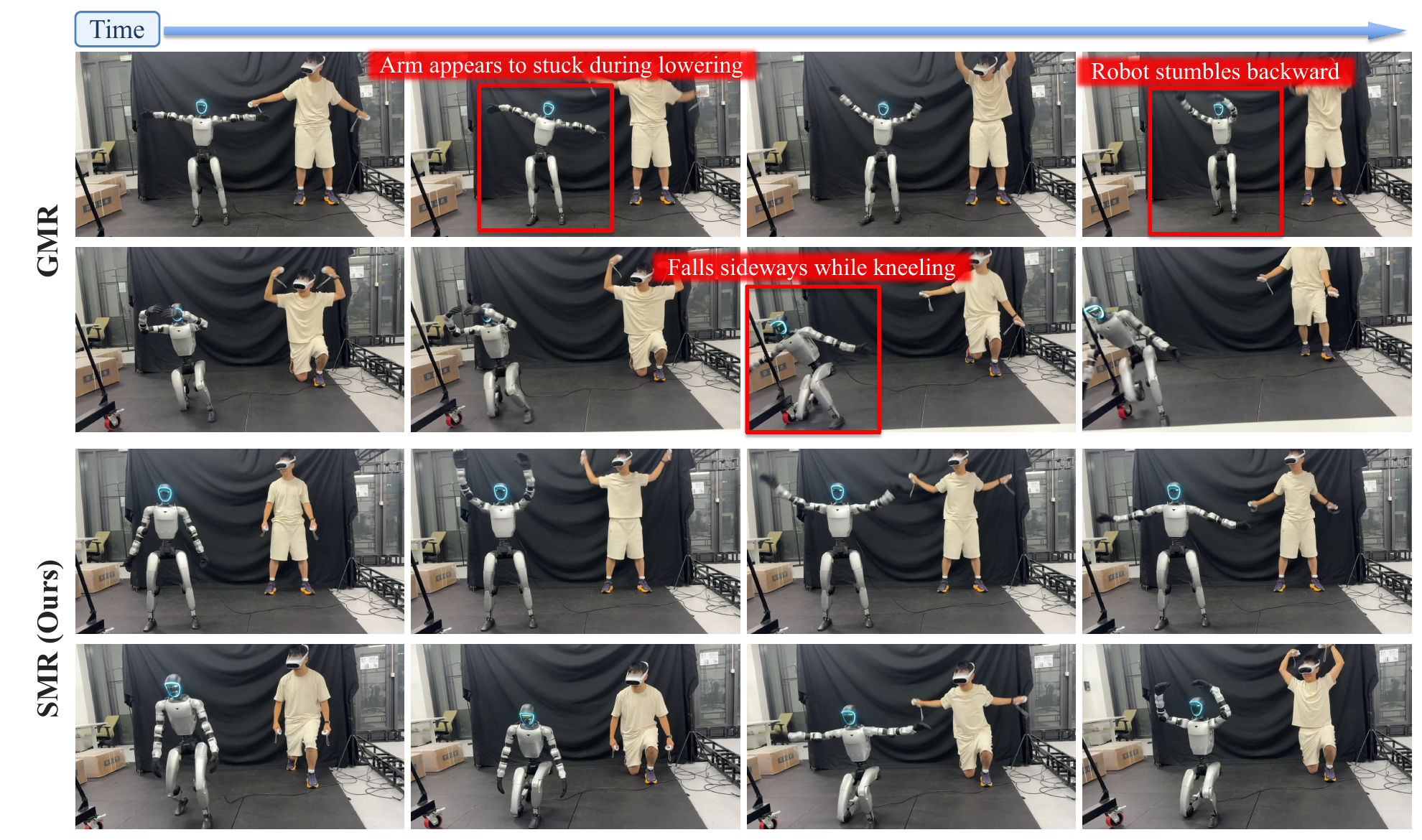}
  \caption{Streaming Unitree G1 teleoperation driven by GMR (top two rows)
  and SMR (bottom two rows) through MimicLite. Each row shows four frames from an
  independent recording in temporal order, covering arm motion or kneeling.}
  \label{fig:real_robot_teleop}
\end{figure*}

%% file: files/5-conclusion.tex
\section{Conclusion}

SMR learns to generate robot references from a live stream of human poses.
Physics experts track robot-aware landmarks in simulation, letting contact and
dynamics shape the trajectories that supervise a causal retargeter. The model
then generates each reference from the human poses observed so far. Direct
landmark tracking improves expert motion fidelity. In G1 simulation, SMR
improves completion and tracking accuracy over GMR and NMR under a shared
frozen MimicLite controller. We also demonstrate streaming teleoperation on
hardware.

\section{Limitations and future work}
We train and evaluate SMR on a single robot, the Unitree G1. For another
humanoid, the semantic mapping can be reused across motions once its neutral
pose and landmark correspondences are specified. The experts and retargeter
are trained for the target robot, and expert training requires substantial GPU
resources. Future work will extend SMR to more robot platforms and accelerate
expert training.

%% file: files/9-appendix.tex

\section{Semantic-conditioned physics experts}
\label{sec:supp_physics_experts}

This section specifies the construction of the robot-aware semantic command,
the offline expert population, and the accepted simulator rollouts used as
student supervision.

\subsection{Robot-aware semantic command details}
\label{sec:supp_semantic_command}

We construct the semantic command $c_t=\Psi_{\mathcal R}(H_t)$ from a
canonical SMPL-H motion $H_t$ at 50~Hz. For source archives that store
translation in leg-length units, we recover meters using the reference
leg length recorded with the input. Forward kinematics uses the neutral
SMPL-H template, and we rotate the resulting poses from source Y-up to
robot Z-up. The extracted skeleton contains 17 joints, including
both toes.  The toes are used only to determine the vertical
alignment and are excluded from the landmark set $\mathcal K$.  The 15
landmarks in $\mathcal K$ are the pelvis, neck, head, and the hips, knees,
ankles, shoulders, elbows, and wrists on both sides.  Each landmark has a
position, orientation, linear velocity, and angular velocity.

\suppressfloats[t]
\input{tables/landmark-set}

\paragraph{Morphology-aware transport.}
All transport parameters are fixed by the neutral SMPL-H rest pose and the
Unitree G1 neutral configuration.  Let $\ell_{\mathcal R}$ and $\ell_{\mathrm S}$
be the sums of the hip--knee and knee--ankle rest lengths over both legs of the
Unitree G1 and SMPL-H skeleton, respectively.  We scale the three coordinates of the
human root by
$s_{\mathcal R}=\ell_{\mathcal R}/\ell_{\mathrm S}$ and anchor this scaling at
the world origin.  A constant vertical offset then aligns the lowest toe of
the scaled human rest pose with the corresponding robot rest pose.  Cropping
a sequence therefore does not change the transported pose at an unchanged
timestamp.  The remaining landmarks are assembled along the human segment
directions with lengths from the robot rest pose rather than by pointwise
scaling.  The neck is attached to a fixed waist anchor, and the head is placed
using the neutral Unitree G1 head geometry.  For every landmark, including the
root, we map the SMPL orientation into a reference frame fixed at the Unitree G1
neutral configuration.  This transport produces task-space landmark targets only.  It
performs no inverse-kinematics solve and produces no robot joint trajectory.
Setting up this transport for another robot requires specifying its neutral
configuration and corresponding landmarks once. The resulting mapping is
then applied automatically to each frame. Different body layouts may require
revising the landmark set.

\paragraph{Command fields.}
For each landmark, the command contains the transported position and
orientation and their linear and angular velocities.  All four fields are
computed after transport and archived in the world frame at 50~Hz.  The
velocities are finite differences of the transported quantities and are used
by the expert critic and tracking rewards, not as additional student inputs.
Section~\ref{sec:supp_expert_details} specifies the frame conventions under
which the expert observes this command.

\subsection{Motion grouping and expert assignment}
\label{sec:supp_motion_grouping}

We partition the heterogeneous motion corpus into offline training pools
before fitting the physics experts.  This subsection details the Harmonic
Motion Embedding (HME) grouping summarized in
Section~\ref{sec:physics_expert}.  The grouping input is the robot-aware
semantic command sequence at 50~Hz. We encode 83 features per frame using
root-relative landmark positions, pelvis height and tilt, five segment
orientations, and four root-motion channels. Motions shorter than the
101-frame, two-second window are padded by holding the terminal pose and
zeroing the padded velocity channels.

\paragraph{Periodic autoencoder.}
A Periodic Autoencoder (PAE)~\citep{starke2022deepphase} encodes each
101-frame window with a convolutional encoder and 64 phase channels.  A
Fourier decomposition of each phase channel provides a harmonic amplitude and
a power-weighted frequency.  We train one PAE per clustered source with
reconstruction error as the only objective, using 50{,}000 AdamW steps,
learning rate $3\times10^{-4}$, weight decay $10^{-4}$, batch size 512, 16
windows per motion, a 2\% validation split, and random seed 0.  We slide the
same window with a stride of 0.2~s over each complete motion.  Concatenating
the mean and standard deviation of the 64 amplitudes and 64 frequencies over
these windows yields one 256-dimensional descriptor per motion.  We do not
segment a motion into separate descriptors.

\paragraph{Clustering and expert assignment.}
We standardize each descriptor dimension and apply MiniBatch K-means
separately within each large source, with $K=256$ for BONES-SEED and $K=32$
for 100STYLE.  Each nonempty cluster forms the training pool of one
cluster expert.  The smaller specialized sources bypass clustering and form
one single-motion expert per motion. Table~\ref{tab:expert_assignment} summarizes
the updated Unitree G1 roster of 428 experts for 145{,}933 candidate motions,
including a 40-motion LAFAN1 subset. Cluster and expert identities are used
only to assign the offline training pools.

\subsection{Physics expert training}
\label{sec:supp_expert_details}

\input{figures/real-robot-teleop-tasks}

We train each expert as a closed-loop PPO policy in the target-robot
simulator.  The semantic command window contains eight samples at 50~Hz:
the current sample, three samples from the preceding 160~ms, and four samples
from the following 80~ms.  The proprioceptive history uses offsets
$\{0,1,2,3,4,8,16\}$ control steps in the past.  Within the command window,
the actor observes only position and orientation fields.  Its full observation
also includes the sparse proprioceptive history and the three most recent
policy actions (\cref{tab:expert_observations}).  Landmark velocities are
reserved for the privileged critic and for tracking rewards.  The policy
outputs a 29-dimensional joint-position
target at 50~Hz.  Each of the eight tracking terms uses the kernel
$\exp(-\bar e/\sigma)$ with its own width.  Position tracking covers the
root and every landmark of $\mathcal K$ except the head, and the five
selected orientation bodies are the torso, both wrists, and both ankles.
Root displacement and orientation residuals are encoded in the current
  robot heading-local frame for the actor, while root tracking errors are
  evaluated in the world frame and body-landmark errors in the current
  heading-local frame.  A curriculum changes the tracking widths from 1.25 to
0.5 times their base values during the first 1{,}050 iterations, while the
weights in \cref{tab:expert_reward_weights} remain fixed.  An episode
terminates when a root or landmark error exceeds its threshold for consecutive
control steps or when the motion ends.

\input{tables/expert-observation-reward}

\paragraph{Optimization, episodes, and reset states.}
The actor and critic are MLPs with Mish activations and running observation
normalization.  The actor has three 1{,}024-unit hidden layers, and the critic
has hidden layers of 1{,}024, 512, and 256 units.  PPO uses a clip ratio of
0.2, entropy coefficient 0.012, AdamW learning rate $5\times10^{-5}$ with an
adaptive KL schedule, discount $0.99$, and $\lambda=0.95$.  We train with
16{,}384 parallel environments for 1{,}500 iterations for single-motion
experts and 4{,}000 iterations for cluster experts.  After a failure, an
episode restarts with probability 0.8 from 25--125 frames earlier,
with the start frame selected by an adaptive difficulty sampler.  The initial
robot state is obtained from offline differential IK evaluated at 50~Hz.
Each solve prioritizes landmark positions, adds low-weight orientation
targets and a weak prior toward the previous frame's posture, and enforces
joint limits and a 30~rad/s bound on the per-frame joint change. The solve
uses the semantic sequence and the G1 model without contact optimization.
PPO loads the resulting pose and velocity at the sampled reset frame and
tracks the semantic commands throughout the episode.

\subsection{Expert rollout supervision}
\label{sec:supp_acceptance}

After training, each expert deterministically replays every assigned motion
once at 50~Hz with its mean policy, starting from the same offline IK reset
state used during training.  The recorded simulator state at frame $t$ is
paired with the semantic command at frame $t$, which is the frame used by the
tracking reward.  A rollout is accepted only if it reaches the final frame
without termination and passes every check in \Cref{tab:acceptance_gates}.
The updated release accepts 89{,}686 of 145{,}933 candidates. We assign
accepted trajectories by a deterministic hash of their original motion
group, with target proportions of 90/5/5. Chunks and mirrored variants of
the same source group share a split. This yields 80{,}728 training,
4{,}513 validation, and 4{,}445 test trajectories. The experts are trained
before this split, so it assesses student distillation on held-out groups.

The frozen reference-quality benchmark in Tab.~\ref{tab:reference_quality} uses
the earlier 465-expert roster of 145{,}970 candidate motions. That release
accepted 105{,}634 semantic-expert trajectories and 106{,}019 GMR-conditioned
expert trajectories.

\input{tables/acceptance-gates}

\section{Neural retargeting}
\label{sec:supp_neural_retargeter}

This section specifies the causal student's input and output contracts,
streaming state, and training recipe.

\paragraph{Input and output contract.}
At time $t$, the input combines the 79-D human pose and 6-D human-root
twist defined in Sec.~\ref{sec:neural_retargeting}. The twist is a backward difference
computed at the native observation rate and is zero at the first frame of
each session.  The student receives no simulator state, expert action or
identity, or future human observation.  The output is the 35-D robot motion
reference $\widehat m_t=(\widehat q_t,\widehat{\Delta p}_t,
\widehat{\Delta r}_t)$: 29 Unitree G1 joint positions, a 3-D pelvis translation
offset, and a 3-D relative pelvis rotation.  The two offsets are expressed
in the current human-pelvis frame and are composed with the current human
pelvis pose to reconstruct the robot root.  This representation avoids
integrating a predicted root velocity, but the reconstructed root can still
inherit human-root drift, heading noise, and ground-height error.  The
student produces a motion reference, not a torque or action.  The downstream
whole-body controller executes this reference.

\paragraph{Architecture and streaming.}
The deployed model has four pre-normalized Transformer blocks of width
512 and 16 attention heads.  Each head has dimension 32, and the
feed-forward layer is four times the model width with GELU activations.
Rotary embeddings use base 10{,}000, dropout is disabled, and the model
  has approximately 12.7~M parameters.  RoPE positions are scaled by $50/f$
at input rate $f$, so the same elapsed time receives the same rotary
offset across rates. The mask gives each layer access to the current
token and up to 127 preceding tokens. During
streaming inference, each layer maintains a bounded ring KV cache and
processes one arriving frame per call.  The cache is reset at a session
boundary.  Input normalization is folded into the deployed graph, and
  outputs are reported in physical units.  Strict causality describes this
  retargeter interface.  The downstream WBC may use an 80-ms future-reference
  buffer.  This buffer is not retargeter lookahead and is reported separately.

\paragraph{Training recipe.}
The updated checkpoint uses 50{,}000 AdamW steps with a batch of 4{,}096
windows and a cosine learning-rate schedule from $3\times10^{-4}$ to
$3\times10^{-5}$. The earlier Student-Final benchmark uses 15{,}000 steps
and batch size 32. Each window contains up to 256 frames. For offline
resampling, positions are linearly interpolated and orientations use
shortest-arc interpolation at the common human and expert timestamps.
An input window is augmented with probability 0.5. The augmentation
combines temporally correlated 2-mm translation and landmark-position
noise, 0.01-rad selected orientation noise, and one- or two-frame causal
holds initiated with per-frame probability 0.06. Root twist is recomputed
from the augmented input, and the offset label is recomputed relative to
the augmented pelvis using the clean expert pose at the original timestamp.

For the velocity loss, we compare joint and pelvis linear finite
differences and consecutive pelvis relative rotation matrices. The
penetration loss penalizes predicted foot collision geometry more than
2~mm below the floor. The slip loss matches horizontal foot velocity
over intervals where the expert foot moves below 0.15~m/s and its lowest
geometry point is below 4~cm at both endpoints.

\section{Real-world household teleoperation}
\label{sec:supp_real_robot_teleop_tasks}

Figure~\ref{fig:supp_real_robot_teleop_tasks} shows three household
teleoperation trials with a Unitree G1. From top to bottom, the robot moves
a pillow on a bed, walks while carrying a bag, and handles a cloth at a
washing machine.
These trials show that SMR can support teleoperated data collection for
real-world household tasks.

%% file: tables/landmark-set.tex
\begin{table}[t]
  \centering
  \small
  \caption{The predefined landmark set $\mathcal K$.}
  \label{tab:landmark_set}
  \begin{tabular*}{\columnwidth}{@{\extracolsep{\fill}}ccc@{}}
    \toprule
    Trunk & Leg (per side) & Arm (per side) \\
    \midrule
    Pelvis, neck, head & Hip, knee, ankle & Shoulder, elbow, wrist \\
    \bottomrule
  \end{tabular*}
\end{table}

%% file: figures/real-robot-teleop-tasks.tex
\begin{figure*}[t]
  \centering
  \includegraphics[width=\linewidth]{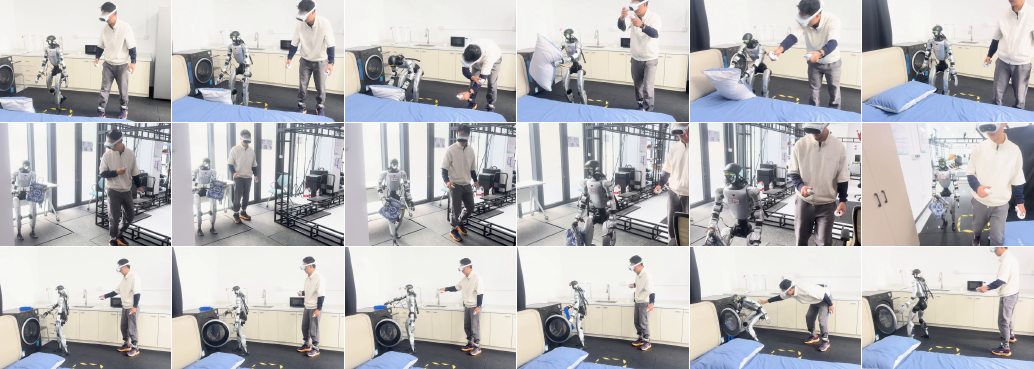}
  \caption{Unitree G1 teleoperation in real-world household settings.
  Each row shows six frames from one recording, with time progressing
  from left to right.}
  \label{fig:supp_real_robot_teleop_tasks}
\end{figure*}

%% file: tables/expert-observation-reward.tex
\begin{table}[t]
  \centering
  \footnotesize
  \renewcommand{\arraystretch}{1.08}
  \caption{Actor observations for semantic-conditioned experts.}
  \label{tab:expert_observations}
  \begin{tabular*}{\columnwidth}{@{\extracolsep{\fill}}lc@{}}
    \toprule
    Actor observation & Dim. \\
    \midrule
    \multicolumn{2}{@{}l}{\textit{Semantic command window (8 samples)}} \\
    Root displacement & $8\times3=24$ \\
    Root orientation & $8\times6=48$ \\
    Tracked body position & $8\times13\times3=312$ \\
    Selected body orientation & $8\times5\times6=240$ \\
    \midrule
    \multicolumn{2}{@{}l}{\textit{Robot proprioception history (7 samples)}} \\
    Root angular velocity & $7\times3=21$ \\
    Projected gravity & $7\times3=21$ \\
    Joint position offset & $7\times29=203$ \\
    Joint velocity & $7\times29=203$ \\
    \midrule
    \multicolumn{2}{@{}l}{\textit{Policy action history (3 actions)}} \\
    Joint-position action & $3\times29=87$ \\
    \midrule
    Total & 1159 \\
    \bottomrule
  \end{tabular*}
\end{table}

\begin{table}[t]
  \centering
  \footnotesize
  \renewcommand{\arraystretch}{1.08}
  \caption{Reward weights for semantic-conditioned expert training.}
  \label{tab:expert_reward_weights}
  \begin{tabular*}{\columnwidth}{@{\extracolsep{\fill}}lc@{}}
    \toprule
    Expert reward & Weight \\
    \midrule
    \multicolumn{2}{@{}l}{\textit{Command tracking}} \\
    Root position & 1.0 \\
    Root orientation & 0.5 \\
    Root linear velocity & 0.5 \\
    Root angular velocity & 0.5 \\
    Body position & 1.0 \\
    Body orientation & 0.5 \\
    Body linear velocity & 0.5 \\
    Body angular velocity & 0.25 \\
    \midrule
    \multicolumn{2}{@{}l}{\textit{Regularization}} \\
    Action rate & 0.5 \\
    Joint velocity & $10^{-4}$ \\
    Soft joint limits & 10.0 \\
    Self-collision & $-1.0$ \\
    Survival & 5.0 \\
    \bottomrule
  \end{tabular*}
\end{table}

%% file: tables/acceptance-gates.tex
\begin{table}[t]
  \centering
  \footnotesize
  \caption{Label-production acceptance gates.}
  \label{tab:acceptance_gates}
  \begin{tabular}{@{}p{.35\columnwidth}@{\hspace{5pt}}c@{}}
    \toprule
    Check & Threshold \\
    \midrule
    Completion & full target length, no termination \\
    Joint limits & \parbox[t]{\dimexpr.65\columnwidth-5pt\relax}{\centering
      violation fraction $\le 0.001$,\\ margin $\ge -0.02$~rad} \\
    Joint speed & $\le 40$~rad/s \\
    Effort saturation & fraction $\le 0.05$ at $0.95\times$ limit \\
    Self-collision & contact fraction $\le 0.01$ above 1~N \\
    Root position & p95 error $\le 0.20$~m \\
    Body position & max p95 error $\le 0.20$~m \\
    Root orientation & p95 error $\le 40^\circ$ \\
    Selected orientations & p95 error $\le 40^\circ$ \\
    Persistent orientation & runs above $60^\circ$ shorter than 25 frames \\
    \bottomrule
  \end{tabular}
\end{table}

%% file: files/7-ethics-statement.tex
\subsection*{Ethics statement}

We use and redistribute source motions and derived assets under their data
and model licenses. We collect, store, and share potentially identifiable
recordings in accordance with participant consent and privacy protections.
Physical deployment requires a separate risk assessment, supervised testing,
and safeguards against falls and collisions.

%% file: files/8-reproducibility-statement.tex
\subsection*{Reproducibility statement}

Appendix~\ref{sec:supp_physics_experts} details semantic targets, expert
training, and rollout acceptance. Appendix~\ref{sec:supp_neural_retargeter}
covers student training and streaming inference.
Section~\ref{sec:experiments} describes the evaluation setup.